\documentclass[letterpaper, 10 pt, conference]{ieeeconf}  

\IEEEoverridecommandlockouts                              

\usepackage{amssymb}  
\usepackage{makecell}
\usepackage{cite}
\usepackage{ulem} 
\usepackage{balance}
\usepackage{graphicx} 
\usepackage{stfloats}
\usepackage{hyperref}   
\usepackage{float}
\usepackage[ruled]{algorithm2e}
\usepackage{caption}
\usepackage{multirow} 
\usepackage{booktabs}
\usepackage{indentfirst}
\usepackage{subcaption}
\usepackage{color}  
\usepackage{lipsum} 
\usepackage{amsmath,esint} 
\usepackage{cuted}
\usepackage{CJKutf8}
\usepackage{verbatim}
\usepackage{bm}
\usepackage{setspace}
\usepackage{array,caption}
\usepackage{fancyhdr}
\usepackage[labelfont=bf]{caption}
\usepackage{ulem}
\usepackage{color}
\usepackage{xcolor}
\usepackage{amsfonts}
\usepackage{tabularray}
\usepackage[table,xcdraw]{xcolor}

\hypersetup{
	colorlinks=true,
	linkcolor=blue,
	citecolor=blue,
	urlcolor=blue,
}

\title{\LARGE \bf
P-FABRIK: A General Intuitive and Robust Inverse Kinematics Method for Parallel Mechanisms Using FABRIK Approach
}

\author{Daqian Cao, Quan Yuan, and Weibang Bai$^\ast$ 
\thanks{
This work is supported by the Shanghai Pujiang Program under grant 23PJ1408500, by the Shanghai Frontiers Science Center of Human-centered Artificial Intelligence (ShangHAI), MoE Key Laboratory of Intelligent Perception and Human-Machine Collaboration (KLIP-HuMaCo). The experiments of this work were supported by the Core Facility Platform of Computer Science and Communication, SIST, ShanghaiTech University.
Corresponding author: Weibang Bai \textit{(wbbai@shanghaitech.edu.cn)}.
}
\thanks{Daqian Cao, Quan Yuan, and Weibang Bai are with the ShanghaiTech Automation and Robotics (STAR) Center, School of Information Science and Technology, ShanghaiTech University, Shanghai, 201210, China.}
}

\begin{document}

\maketitle
\thispagestyle{empty}
\pagestyle{empty}

\begin{abstract}
Traditional geometric inverse kinematics methods for parallel mechanisms rely on specific spatial geometry constraints. However, their application to redundant parallel mechanisms is challenged due to the increased constraint complexity. Moreover, it will output no solutions and cause unpredictable control problems when the target pose lies outside its workspace.
To tackle these challenging issues, this work proposes P-FABRIK, a general, intuitive, and robust inverse kinematics method to find one feasible solution for diverse parallel mechanisms based on the FABRIK algorithm. 
By decomposing the general parallel mechanism into multiple serial sub-chains using a new topological decomposition strategy, the end targets of each sub-chain can be subsequently revised to calculate the inverse kinematics solutions iteratively. Multiple case studies involving planar, standard, and redundant parallel mechanisms demonstrated the proposed method’s generality across diverse parallel mechanisms. Furthermore, numerical simulation studies verified its efficacy and computational efficiency, as well as its robustness ability to handle out-of-workspace targets.

\end{abstract}

\section{Introduction}
Compared to serial robots, parallel robots perform better in applications demanding high load capacity and high precision \cite{Patel2012}. These advantages make parallel mechanisms widely used in multiple regions. Such as machining, pipeline operating, satellite tracking, and surgical operating \cite{bruzzone2002modelling, Connolly2007, Jones2003, Li2007}. To control the parallel robots, inverse kinematics (IK) analysis of parallel mechanisms is a fundamental requirement.  

Geometric IK methods are widely used to analyze the parallel mechanisms. Changela et al. \cite{changela2012geometric} use the geometric approach to solve the IK solution of a 3-\underline{P}SU parallel kinematic manipulator. Bueno et al. \cite{bueno2021geometric} details an accessible geometric derivation of the IK of a parallel robotic linkage known as the Canfield joint. Pashkevich et al. \cite{pashkevich2004orthoglide} analyze the IK of Orthoglide, a three-DOF parallel mechanism, with the geometric method. Williams et al. \cite{williams1997inverse} present algebraic inverse position and velocity kinematics solutions for a broad class of 3-degrees-of-freedom (3-DoFs) planar in-parallel-actuated manipulators using a geometric approach. However, traditional geometric IK methods for parallel mechanisms rely on specific spatial geometry constraints. The complexity of these constraints increases when applying the methods to redundant parallel mechanisms, which often have complex sub-chain configurations. Moreover, traditional geometric IK methods will output no solutions and cause unpredictable control problems when the target pose of the mechanism lies outside its workspace.

To tackle these challenging problems, we propose P-FABRIK, a general, intuitive, and robust inverse kinematics method to find one feasible solution for diverse parallel mechanisms based on the Forward and Backward Reaching Inverse Kinematics (FABRIK) algorithm \cite{aristidou_fabrik_2011}.

The FABRIK algorithm was originally formulated for computer graphics. Its inherent generality, intuitiveness, and robustness have subsequently enabled the successful migration into robotics research domains \cite{kolpashchikov2018fabrik, moreno2022adaptation}. 
Santos et al. \cite{Santos2020} proposed an algorithm based on FABRIK with modified base constraints for mobile manipulator systems. Santos et al. \cite{Santos2021} developed a FABRIK-based IK solver for robotic manipulators by incorporating additional restrictions to model the 1-DoF joints. Zhang et al. \cite{Zhang2018} introduced a curvature-aware extension for FABRIK to solve the IK of continuum robots, etc. 
However, none of those methods involved solving the IK for parallel mechanisms due to the problematic closed kinematic topology issue. In this paper, we proposed a topological decomposition strategy (TDS) and an adaptive target projection (ATP) method within the P-FABRIK to enlarge the application of the FABRIK algorithm.

The main contributions of this work are listed as follows:
\begin{itemize}
    \item [1.] P-FABRIK is proposed as a general, intuitive, and robust kinematic method for parallel mechanisms using the FABRIK approach, during which TDS and ATP methods are proposed.
    \item [2.] The detailed iteration algorithm and various verification case studies of the proposed P-FABRIK are provided, demonstrating its generality.
    \item [3.] Numerical simulations are performed to evaluate the efficacy, robustness, and computational efficiency of the proposed P-FABRIK algorithm.
\end{itemize}

\section{P-FABRIK: FABRIK FOR PARALLEL MECHANISMS}\label{sec:p-fabrik}
\subsection{Background of FABRIK}\label{sec:fabrik}

FABRIK is a heuristic method that directly locates the joint's position without using rotational matrices. The basic iteration process and the termination criterion of this algorithm are demonstrated in this section.

\begin{figure}[thpb]
      \centering
      \includegraphics[scale=0.35]{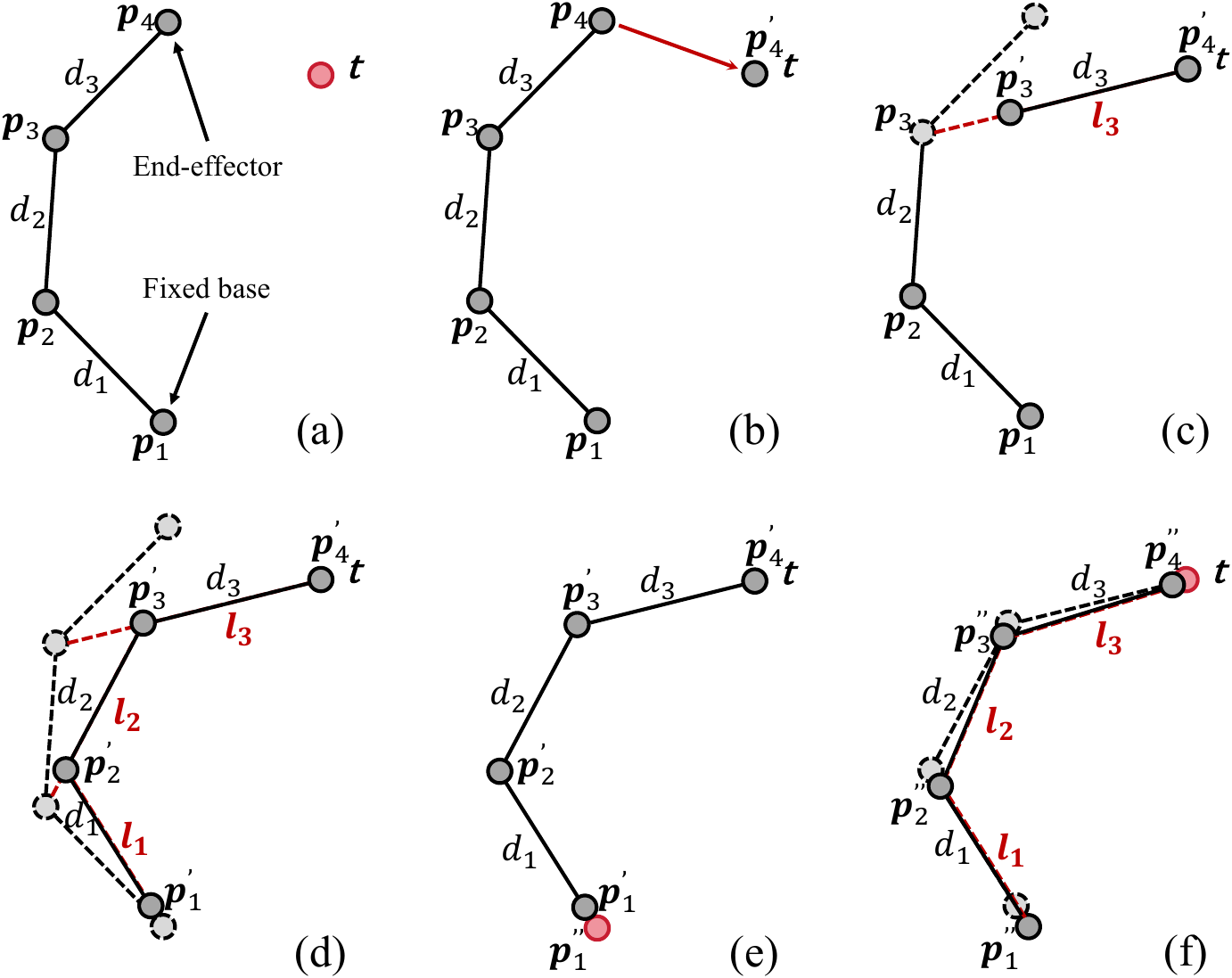}
      \caption{An example of a full iteration of FABRIK for a planar serial mechanism with 4 revolute joints. }
      \label{fig:fabrik-example}
   \end{figure}

The FABRIK algorithm implements an initial reachability assessment of the target by summing up all link lengths of the mechanism. For reachable targets, the iterative process operates through two distinct computational stages: the forward-reaching stage and the backward-reaching stage. Fig. \ref{fig:fabrik-example} illustrates a full iteration of FABRIK for a 4-joint planar serial mechanism, where $p_{1}$, $p_{2}$, $p_{3}$, and $p_{4}$ represent the joints of the mechanism where $p_{1}$ is the fixed base and $p_{4}$ is the end-effector. $d_{1}$, $d_{2}$, and $d_{3}$ represent the mechanism's link lengths, and $t$ denotes the target point. It should be noted that any point denoted by a bold symbol represents its position vector.

The forward-reaching stage starts from the end effector and adjusts joint positions along the kinematic chain toward the base joint iteratively. As shown in Fig. \ref{fig:fabrik-example}(b), $\boldsymbol{p}_{4}$ is moved to $\boldsymbol{t}$. The new position is denoted as $\boldsymbol{p}^{\prime}_{4}$. $\boldsymbol{p}^{\prime}_{3}$ should be located on the line $\boldsymbol{l}_{3}$ that passes through $\boldsymbol{p}_{3}$ and $\boldsymbol{p}^{\prime}_{4}$ with distance $d_{3}$ from $\boldsymbol{p}^{\prime}_{4}$ as shown in Fig. \ref{fig:fabrik-example}(c). This procedural iteration is recursively applied to joints ${p}_{2}$ and ${p}_{1}$. The reconfigured mechanism is illustrated in Fig. \ref{fig:fabrik-example}(d) and $\boldsymbol{p}^{\prime}_{1}$ is different from $\boldsymbol{p}_{1}$ after the forward-reaching stage.

Because FABRIK is designed for mechanisms with a fixed base, the backward-reaching stage is initiated by resetting $\boldsymbol{p}^{\prime}_{1}$ to its original configuration $\boldsymbol{p}^{\prime\prime}_{1}$ as shown in Fig. \ref{fig:fabrik-example}(e). Then, the same steps as the forward-reaching stage are performed outward till the end effector. Fig. \ref{fig:fabrik-example}(f) shows the reconfigured mechanism after one full iteration.

As established in the convergence analysis by Aristidou et al. \cite{Aristidou2015}, the distance between the end effector and the target $||\boldsymbol{t}-\boldsymbol{p}_{4}||$ will converge to zero after sufficient iterations. This process requires alternating between the forward-reaching and backward-reaching stages until the termination criterion below is satisfied,
\begin{equation}\label{eq:termination}
    ||\boldsymbol{t}-\boldsymbol{p}_{4}||\leq E
\end{equation}
where $E$ denotes a preset numerical tolerance threshold. 

\subsection{P-FABRIK: FABRIK for Parallel Mechanisms}\label{sec:p-fabrik}
As introduced previously, the FABRIK algorithm was originally designed to solve the IK for serial mechanisms. In this work, the P-FABRIK algorithm is proposed to give one feasible IK solution for parallel mechanisms using the FABRIK approach. 

\begin{figure}[h!]
      \centering
      \includegraphics[scale=0.5]{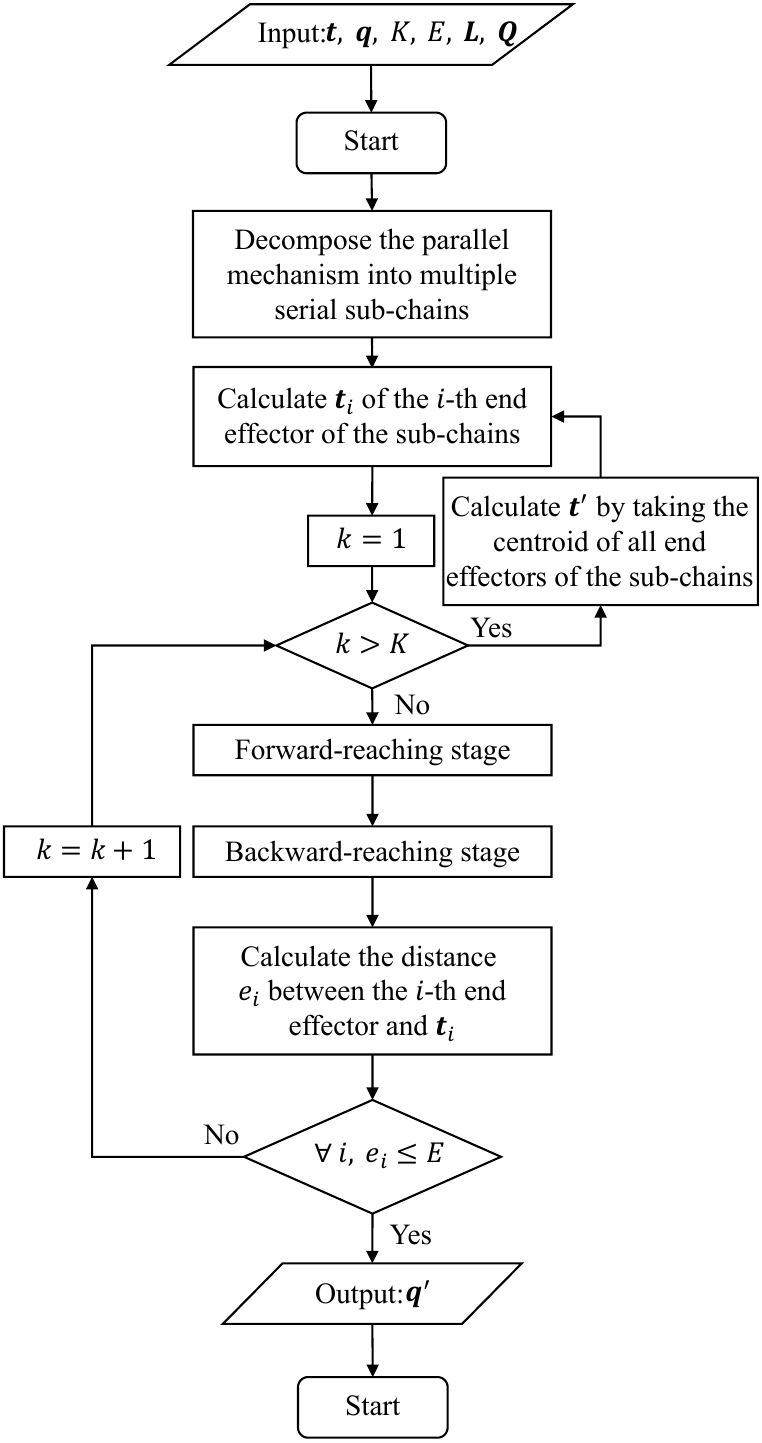}
      \caption{Flowchart of the P-FABRIK algorithm.}
      \label{fig:p-fabrik-flowchart}
\end{figure}

The schematic implementation of the proposed P-FABRIK algorithm is provided in the flowchart in Fig. \ref{fig:p-fabrik-flowchart}, where $\boldsymbol{q}$ and $\boldsymbol{q}^{\prime}$ are the current and the new spatial positions of all joints (except prismatic joints) of the parallel mechanism, respectively. $\boldsymbol{t}$ and $\boldsymbol{t}^{\prime}$ are the original and revised target positions of the parallel mechanism. $\boldsymbol{t}_{i}$ is the target position of the $i$-th end effector of the sub-chains. $K$ is the maximum iteration, and $k$ is the cumulative number of iterations. $\boldsymbol{L}$ represents the length limitations for all links. For simplicity, a general item $\boldsymbol{Q}$ is adopted to denote the angular limitations for each DOF of the revolute, universal, and spherical joints.

A topological decomposition strategy (TDS) is proposed to decompose the parallel mechanism into multiple serial sub-chains. $\boldsymbol{t}_{i}$ is calculated via the inherent geometric relations of the parallel mechanism. $K$ is added to the termination criteria because non-convergence of the algorithm within $K$ iterations suggests that $\boldsymbol{t}$ exceeds the mechanism's workspace.

Therefore, an adaptive target projection (ATP) method is proposed to handle this possible non-convergence caused by the invalid target. The centroid of all end effectors of sub-chains is taken as $\boldsymbol{t}^{\prime}$. Then $\boldsymbol{t}_{i}$ is updated with $\boldsymbol{t}^{\prime}$ for a new round of iteration.

During the forward-reaching and backward-reaching stages, the prismatic joint is considered a link with variable lengths, i.e., the $j$-th prismatic joint is considered the $j$-th link. Its length is checked before the repositioning of other kinds of joints, i.e., revolute, universal, and sphere joints. The repositioning is skipped if
\begin{equation}
    \underline{\boldsymbol{L}_{j}} \leq d_{j} \leq \overline{\boldsymbol{L}_{j}}
\end{equation}
where $d_{j}$ is the length of the $j$-th link, $\underline{\boldsymbol{L}_{j}}$ and $\underline{\boldsymbol{L}_{j}}$ are the lower and upper bounds of the length limitation of the $j$-th link.

The iteration terminates if $k>K$ or $\forall i, e_{i}<E$. The output of P-FABRIK is $\boldsymbol{p}^{\prime}$, by which the actual joint configuration of the parallel mechanism can be calculated. Unless otherwise specified, $K=100$ and $E=10^{-2}$ mm throughout this work.

\section{Case Studies}\label{sec:case-studies}
The verification of the proposed method comprises three distinct phases of IK analyses across multiple parallel mechanisms. The initial phase employs a planar 5-bar mechanism.
The next phase extends to the representative spatial parallel mechanism of the 6-UPS Stewart platform. The final verification phase implements a kinematically redundant parallel mechanism, demonstrating the generality of the proposed method under diverse topological configurations.
\subsection{Planar 5-bar Mechanism}\label{subsec:p-fabrik-planar}
\begin{figure}[h!]
      \centering
      \includegraphics[scale=0.5]{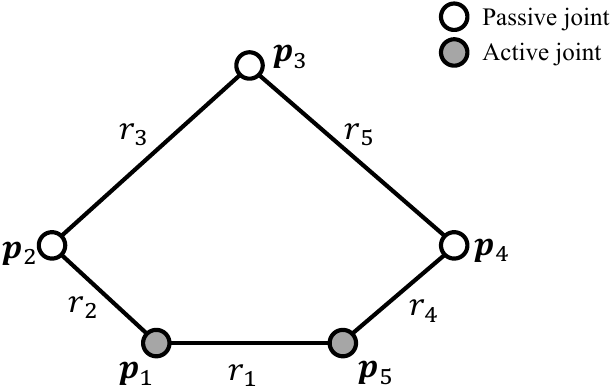}
      \caption{Diagram of a classic planar 5-bar mechanism}
      \label{fig:5-bar-diagram}
\end{figure}

A standard planar 5-bar mechanism usually comprises five revolute (R) joints, with two active joints to establish a 2-DOFs planar motion. A diagram of the classic planar 5-bar mechanism is shown in Fig. \ref{fig:5-bar-diagram}.

In the diagram, $p_{1}$, $p_{2}$, $p_{3}$, $p_{4}$, and $p_{5}$ represent the joints of the planar 5-bar mechanism. $p_{1}$ and $p_{5}$ are active while others are passive. $p_{3}$ is the end effector, while $p_{1}$ and $p_{5}$ are considered as fixed bases. $r_{1}$, $r_{2}$, $r_{3}$, $r_{4}$, and $r_{5}$ are the link lengths of the mechanism. 

As the parallel mechanism has two fixed bases, 
it's divided into two serial sub-chains with TDS. Each sub-chain originates from a distinct fixed base and terminates at the common end effector $p_{3}$. Therefore,
\begin{equation}
    \boldsymbol{t}_{i}=\boldsymbol{t},\ i=1,2
\end{equation}
for each sub-chain $i$.

\begin{figure}[h]
      \centering
      \includegraphics[scale=0.38]{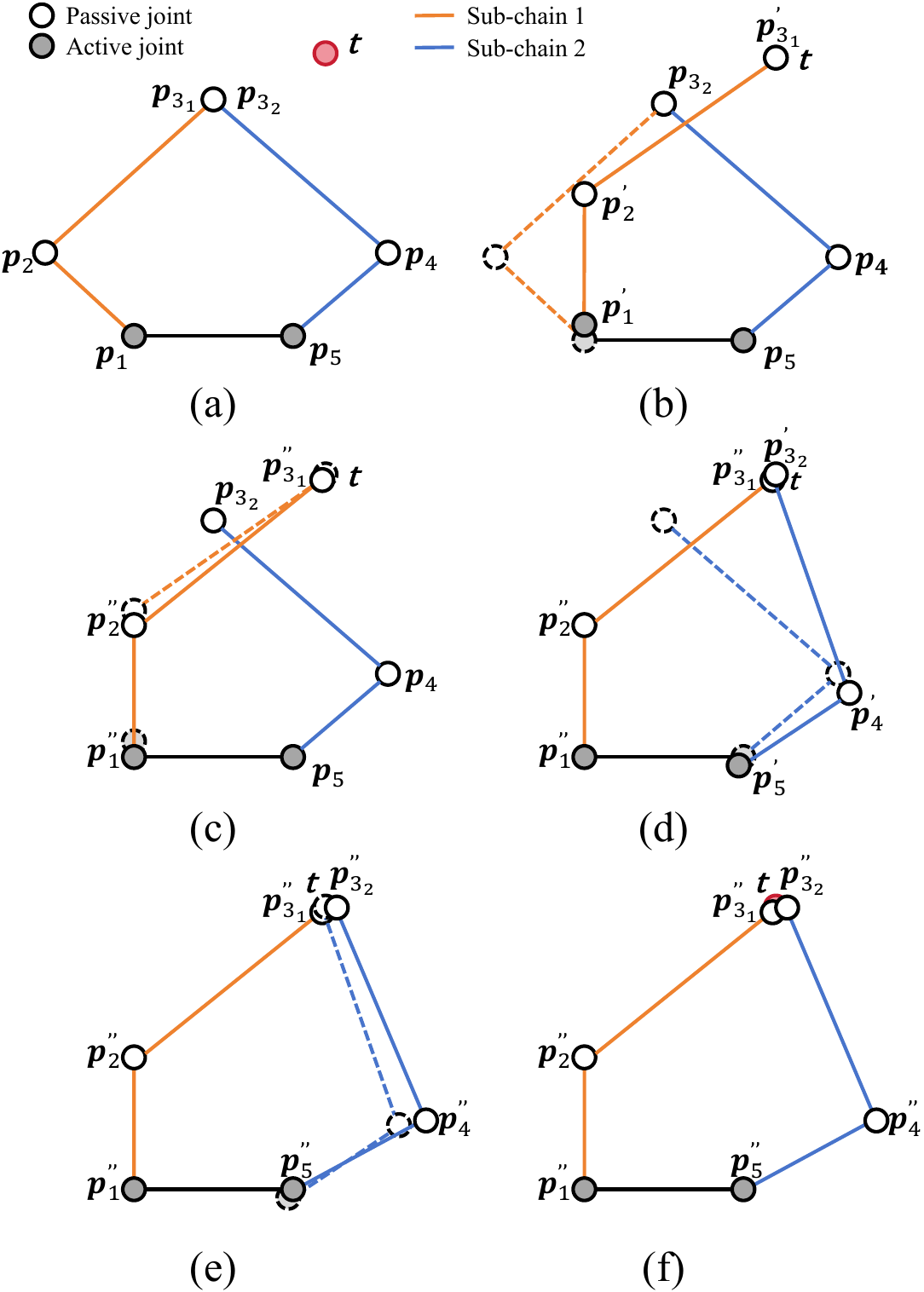}
      \caption{An example of a full iteration of P-FABRIK for a planar 5-bar mechanism. (a) The initial configuration of the 5-bar planar mechanism and the target $\boldsymbol{t}$. The mechanism is decomposed into 2 serial sub-chains. Sub-chain 1's base and end effector are $\boldsymbol{p}_{1}$ and $\boldsymbol{p}_{3_{1}}$ and sub-chain 2's are $\boldsymbol{p}_{5}$ and $\boldsymbol{p}_{3_{2}}$. (b) Apply forward-reaching on sub-chain 1. (c) Apply backward-reaching on sub-chain 1. (d) Apply forward-reaching on sub-chain 2. (e) Apply backward-reaching on sub-chain 2. (e) Reconfigured closed-chain posture after full iteration.}
      \label{fig:5-bar-fabrik}
      \vspace{-3mm}
\end{figure}

Both sub-chains undergo the forward-reaching and backward-reaching stages, as illustrated in Fig. \ref{fig:5-bar-fabrik}. 
The iterations terminate when 
\begin{equation}
    \forall i=1,2,\ ||\boldsymbol{p}_{3_{i}}-\boldsymbol{t}_{i}||\leq E
\end{equation}
or $k>K$.

Supposing the original target position $\boldsymbol{t}$ lies outside the mechanism's workspace, the ATP variation for this case is:
\begin{equation}
    \boldsymbol{t}^{\prime}=\frac{1}{2}(\boldsymbol{p}^{\prime\prime}_{3_{1}}+\boldsymbol{p}^{\prime\prime}_{3_{2}})
\end{equation}
Then, re-initiate the bidirectional reaching stages on both sub-chains using $\boldsymbol{t}^{\prime}$,  constrained by maximum $K$ iterations. 
Meanwhile, joint angle $\theta_{i}$ of joint $\boldsymbol{p}_{i}$ needs to be within the angular bounds, i.e. 
$ \underline{\boldsymbol{Q}_{i}} \leq \theta_{i} \leq \overline{\boldsymbol{Q}_{i}}$,
where $\underline{\boldsymbol{Q}_{i}}$, $\overline{\boldsymbol{Q}_{i}}$ are the lower and upper bound of the joint limitation of joint ${p}_{i}$. 

\subsection{6-UPS Stewart Mechanism}\label{subsec:stewart}
\begin{figure}[htbp]
  \centering
  \begin{subfigure}[t]{0.25\textwidth}
    \includegraphics[width=\textwidth]{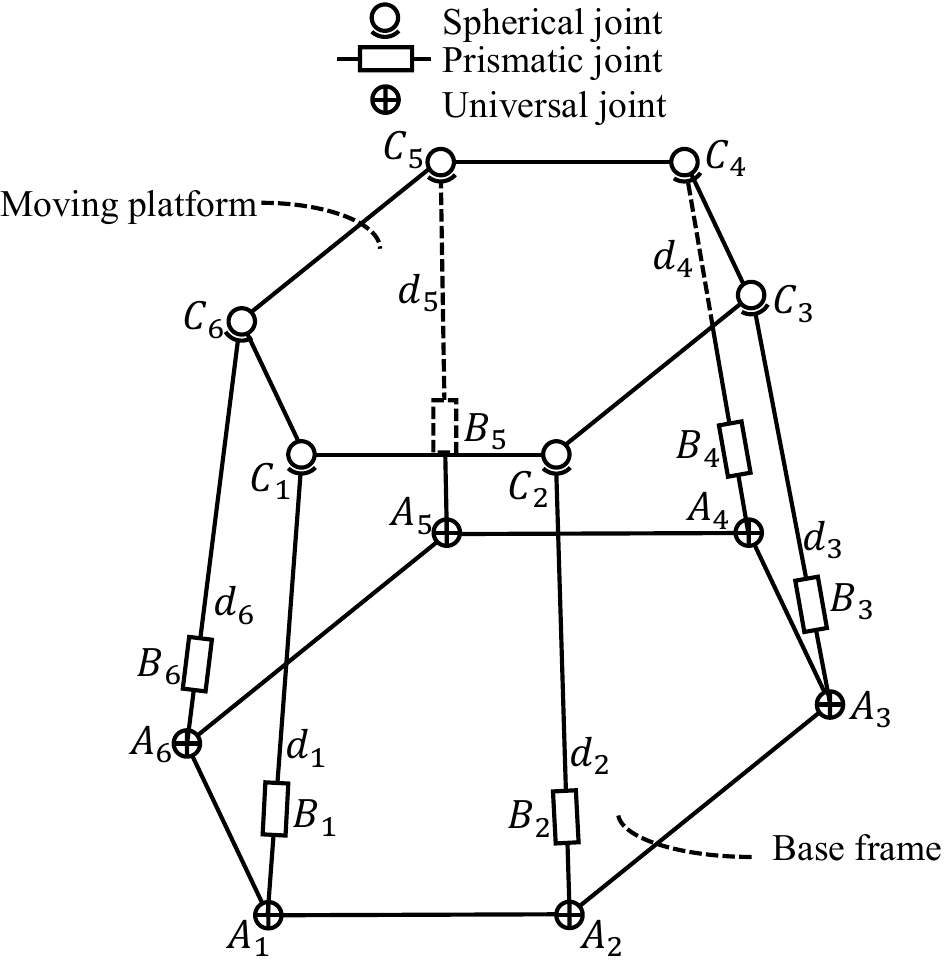}
    \caption{}
  \end{subfigure}
  \begin{subfigure}[t]{0.20\textwidth}
    \includegraphics[width=\textwidth]{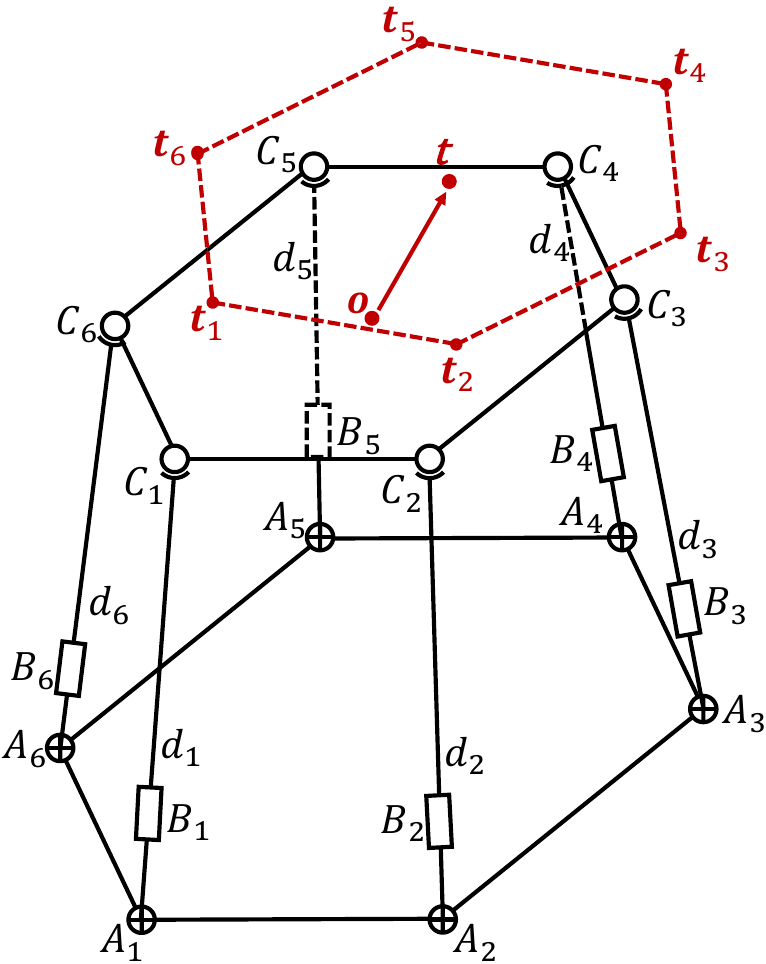}
    \caption{}
  \end{subfigure}
  \caption{(a) Diagram of a 6-UPS Stewart platform. (b) The target $\boldsymbol{t}$ and sub-targets $\boldsymbol{t}_i$ set for the 6-UPS mechanism.}
  \label{fig:stewart-diagram}
\end{figure}
The diagram of the 6-UPS Stewart platform is firstly illustrated in Fig. \ref{fig:stewart-diagram}(a), where $A_{i} \ (i = 1,2, \cdots, 6)$ is the universal joint fixed on the base frame. $B_{i} \ (i = 1, 2, \cdots, 6)$ is the prismatic joint, and $d_{i}$ is the length of $A_{i}C_{i}$. $C_{i} \ (i=1, 2, \cdots, 6)$ is the sphere joint connected to the moving platform. The centroid $\boldsymbol{o}$ of the moving platform, as shown in Fig. \ref{fig:stewart-diagram}(b), is chosen as the reference point. 

The whole parallel mechanism is first decomposed into six serial sub-chains using the previously proposed TDS. Each sub-chain has a fixed base $A_{i}$, a prismatic joint $B_{i}$, and an end effector $C_{i}$. 
For $\boldsymbol{t}$ with a target rotation of the moving platform, the target rotation is first applied to the moving platform with $\boldsymbol{o}$ staying at the same position. Then the transformation matrices from $\boldsymbol{o}$ to $\boldsymbol{C}_{i} \ (i = 1,2, \cdots, 6)$ is denoted as $\boldsymbol{T}_{i} \ (i = 1,2, \cdots, 6)$. The transformation matrices form $\boldsymbol{t}$ to $\boldsymbol{t}_{i} \ (i = 1,2, \cdots, 6)$ is the same as $\boldsymbol{T}_{i}$ to keep the mechanism consistent in structure. Therefore,
\begin{equation}
    \begin{bmatrix}\boldsymbol{t}_{i} \\ 1 \end{bmatrix} =\boldsymbol{T}_{i}\begin{bmatrix}
        \boldsymbol{o}\\
        1
    \end{bmatrix}+ \begin{bmatrix}
        \boldsymbol{t} \\ 1
    \end{bmatrix}, i=1,2,\cdots,6
\end{equation}

 During the forward-reaching and backward-reaching stages, the prismatic joint $B_{i}$ in each sub-chain is considered as a link with a variable length. The value of $d_{i}$ is checked after joint $A_{i}$ or $C_{i}$ is assigned to a new position. If $d_{i}$ is within the length limit of joint $B_{i}$, i.e. $ \underline{\boldsymbol{L}_{i}} \leq d_{i} \leq \overline{\boldsymbol{L}_{i}}$, the algorithm will continue directly without re-positioning the new position of joint $A_{i}$ or $C_{i}$. On the contrary, if $d_{i}$ exceeds the limitation, the new length of joint $B_{i}$ should be
\begin{equation}
    \bar{d_{i}}=\min \left(\max(d_{i}, \ \underline{\boldsymbol{L}_{i}} ), \ \overline{\boldsymbol{L}_{i}} \right)
\end{equation}
Then, joint $A_{i}$ or $C_{i}$ is re-positioned with a distance of $\bar{d_{i}}$ from the previous joint.
 The joint limitations for each DOF of the universal joint and the sphere joint are considered the same. Then, the limited joint angle can be denoted as: 
\begin{equation}
    \bar{\theta_{i}}=\min \left(\max(\theta_{i}, \ \underline{\boldsymbol{Q}_{i}} ), \ \overline{\boldsymbol{Q}_{i}} \right)
\end{equation}
\begin{figure}[h]
  \centering
    \includegraphics[width=0.40\textwidth]{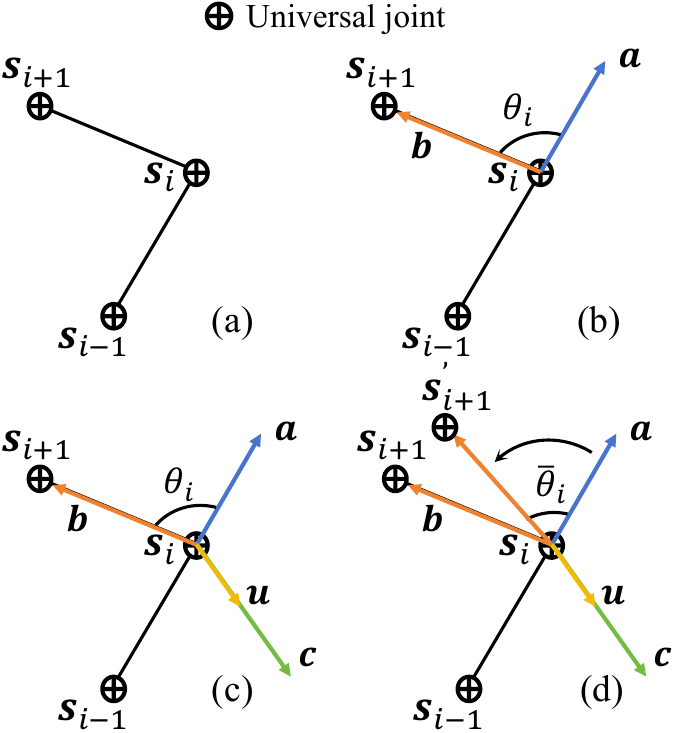}
    \caption{Apply angular limits on $s_{i}$.}
    \label{fig:angular_limits}
  \vspace{-3mm}
\end{figure}
The joint with angular limitations, i.e., the universal and sphere joints,
is denoted as ${s}_{i}$. Fig. \ref{fig:angular_limits} demonstrates the process of applying angular limits on $s_{i}$. As shown in Fig. \ref{fig:angular_limits}(b), a vector $\boldsymbol{a}$ needs to be selected as the angle reference for applying joint limits on ${s}_{i}$. In general, it can be defined as: 
\begin{equation}
    \boldsymbol{a} = \boldsymbol{s}_{i} - \boldsymbol{s}_{i-1}
\end{equation}
As an example, a vector that is perpendicular to the base frame can be chosen as the angle reference of joint $A_{i}$. When limiting the joint ${s}_{i}$, the position of joint ${s}_{i+1}$ may also change.
So another vector $\boldsymbol{b}$ is set to help locating the new position of joint ${s}_{i+1}$:
\begin{equation}
    \boldsymbol{b} = \boldsymbol{s}_{i+1} - \boldsymbol{s}_{i}
\end{equation}
Consequently, a rotation axis $\boldsymbol{u}$ can be obtained by:
\begin{align}
    \boldsymbol{c}&=\boldsymbol{a}\times\boldsymbol{b}\\
    \boldsymbol{u}&=\frac{\boldsymbol{c}}{||\boldsymbol{c}||}
\end{align}
as shown in Fig. \ref{fig:angular_limits}(c).
And the skew-symmetric matrix $\boldsymbol{K}$ for $\boldsymbol{u}=(u_{x},u_{y},u_{z})$ is:
\begin{equation}
    \boldsymbol{K}= \begin{bmatrix}
    0 & -u_{z} & u_{y} \\
    u_{z} & 0 & -u_{x} \\
    -u_{y} & u_{x} & 0
    \end{bmatrix}
\end{equation}
Then the rotation matrix $\boldsymbol{R}$ for calculating the new regulated spatial position of joint $s_{i+1}$ can be obtained by applying Rodrigues’ formula \cite{Dai2015}:
\begin{equation}
    \boldsymbol{R}=\cos{\bar{\theta_{i}}}\times \boldsymbol{I}+\sin{\bar{\theta_{i}}}\times \boldsymbol{K} + (1-\cos{\bar{\theta_{i}}})\times \boldsymbol{uu}^{T}
\end{equation}
The constrained joint position $\boldsymbol{s_{i+1}}$ can be obtained as: 
\begin{equation}
   \boldsymbol{s^{\prime}_{i+1}}=\boldsymbol{R}\frac{\boldsymbol{a}}{||\boldsymbol{a}||}||\boldsymbol{b}||.
\end{equation}
as shown in Fig. \ref{fig:angular_limits}(d).

The termination criteria and the ATP variation for this case are:
\begin{equation}
    \forall i=1,2,\cdots,6,\ ||\boldsymbol{C}_{i}-\boldsymbol{t}_{i}||\leq E\ , 
\end{equation}
or $k> K$, and
\begin{equation}
    \boldsymbol{t}^{\prime}=\frac{1}{6}\sum_{i=1}^{6}\boldsymbol{C}_{i}.
\end{equation}

\subsection{Redundant Parallel mechanism}\label{subsec:nrpm}
\begin{figure}[htbp]
    \centering
    \includegraphics[width=0.42\textwidth]{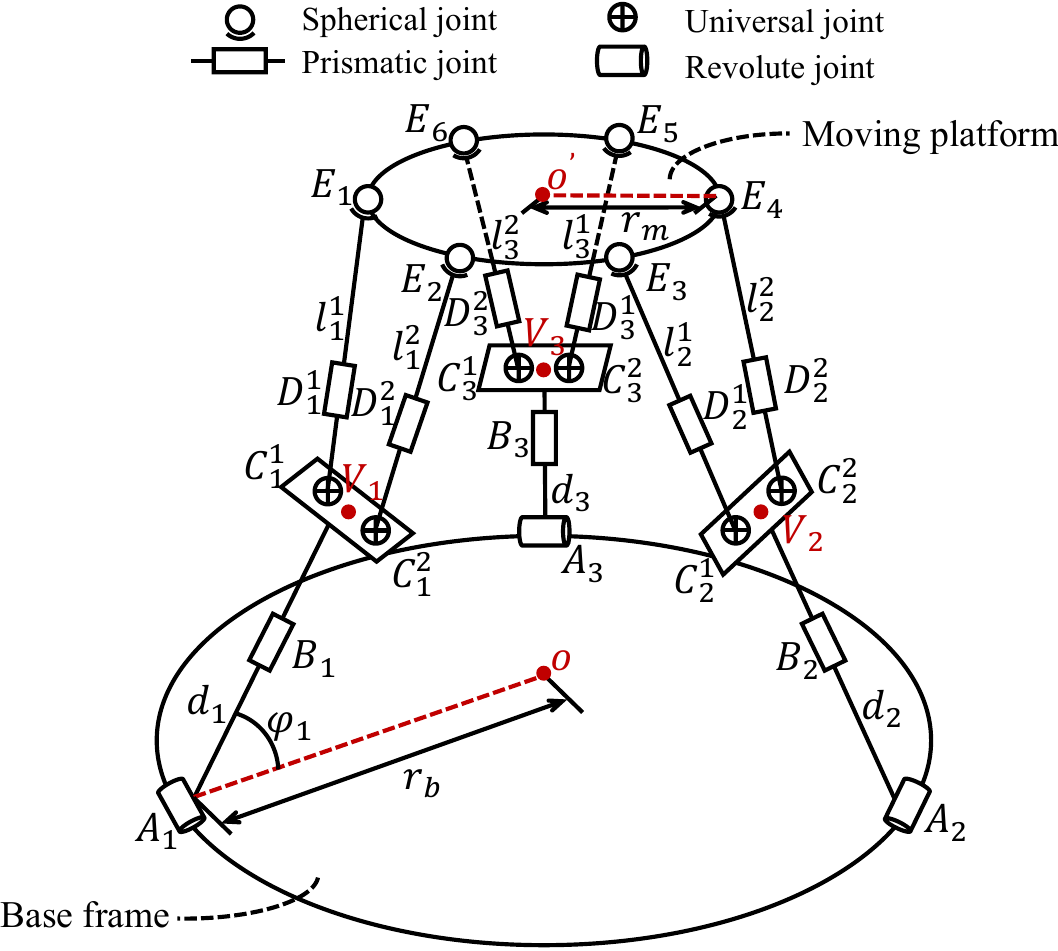}
    \caption{Kinematic diagram of NRPM.}
    \label{fig:nrpm}
\end{figure}

Kinematic redundancy in parallel mechanisms \cite{Marquet, Shayya2013, Zanganeh} is primarily engineered to augment workspace volume and optimize operational dexterity.
Yuan et al. \cite{10801501} developed a novel redundant parallel mechanism (NRPM) to address the inherent limitations of conventional Stewart platform configurations in medical applications. This mechanism is used as a verification platform for the proposed P-FABRIK algorithm. The diagram of NRPM is shown in Fig. \ref{fig:nrpm}. 

As shown in Fig. \ref{fig:nrpm}, $\boldsymbol{o}$ is the centroid of the base frame, whose radius is denoted as $r_{\text{b}}$. $\boldsymbol{o}^{\prime}$ is the centroid of the moving platform, whose radius is denoted as $r_{\text{m}}$. $\varphi_{i}\ (i=1,2,3)$ denotes the angle between $A_iB_i$ and the base frame. The length of $A_iV_i$ is denoted as $d_i$, where $V_i$ is the midpoint of $C_i^1C_i^2$. $l_{i}^{j}\ (i=1,2,3;j=1,2)$ represents the length of $C_i^jE_k\ (k=2(i-1)+j)$. 

By the proposed TDS, NRPM is decomposed into three serial sub-chains. Each sub-chain comprises a revolute joint $A_{i}$ as the fixed base, prismatic joint $B_{i}$, virtual joint $V_{i}$, universal joints $C_{i}^{j}\ (j=1,2)$, prismatic joints $D_{i}^{j}\ (j=1,2)$, and sphere joints $E_{k}$ and $E_{k+1}\ (k=2i-1)$ as end effectors. $V_{i}$ is a fixed joint with 0-DOF, indicates that the relative position between $V_{i}$ and $C_{i}^{j}$, remains the same in the forward-reaching and backward-reaching stages. The methods for calculating $\boldsymbol{t}_i\ (i=1,2,\cdots,6)$ and restricting revolute, universal, and sphere joins are similar to methods presented in Section \ref{subsec:stewart}.

The sub-chains are serial mechanisms with multiple end effectors after the decomposition. Therefore, modifications previously proposed by Aristidou et al. \cite{Aristidou2015} can be applied in the forward-reaching and backward-reaching stages. The termination criteria and the ATP variation for this case are:
\begin{equation}
    \forall i=1,2,\cdots,6,\ ||\boldsymbol{E}_{i}-\boldsymbol{t}_{i}||\leq E\ , 
\end{equation}
or $k> K$, and
\begin{equation}
    \boldsymbol{t}^{\prime}=\frac{1}{6}\sum_{i=1}^{6}\boldsymbol{E}_{i}.
\end{equation}

The generality of the proposed algorithm can be verified now through the three verification phases.

\section{Simulation Verifications}\label{sec:simulations}
In this section, the performance of the proposed P-FABRIK algorithm was demonstrated by conducting simulation experiments. All experiments were programmed by MATLAB and run on a 2.2-GHz i7-13700 CPU platform with Windows 11.

The first experiment evaluated the proposed algorithm's efficacy in a trajectory-tracking scenario.
In this experiment, the planar 5-bar mechanism, 6-UPS Stewart platform, and NRPM were set to track circular trajectories. For both the 6-UPS Stewart platform and NRPM, the target position and orientation of their moving platforms varied along the path.
The IK solutions of these mechanisms obtained from P-FABRIK were verified by the forward kinematics (FK) algorithm of each mechanism. These FK algorithms are based on Newton's method \cite{kelley2003solving}. The initial guess for each FK algorithm is the previous target pose of the mechanism.

The results presented in Tab. \ref{tb:efficacy} show that P-FABRIK could find one feasible solution for diverse parallel mechanisms.

\begin{table}[b]
\caption{Results of verifying the efficacy of P-FABRIK.}
\label{tb:efficacy}
\begin{center}
\begin{tabular}{c|cc}
\hline
Mechanism      & \multicolumn{2}{c}{RMSE}      \\ \hline
Planar 5-bar   & \begin{tabular}[c]{@{}c@{}}Position\\ {[}mm{]}\end{tabular}     & 0.005389 \\ \hline
\multirow{2}{*}{\begin{tabular}[c]{@{}c@{}}6-UPS\\ Stewart platform\end{tabular}} & \begin{tabular}[c]{@{}c@{}}Position\\ {[}mm{]}\end{tabular}     & 0.002887 \\
                                                                                  & \begin{tabular}[c]{@{}c@{}}Orientation\\ {[}deg{]}\end{tabular} & 0.000417 \\ \hline
\multirow{2}{*}{NPRM}                                                             & \begin{tabular}[c]{@{}c@{}}Position\\ {[}mm{]}\end{tabular}     & 0.002504 \\
                                                                                  & \begin{tabular}[c]{@{}c@{}}Orientation\\ {[}deg{]}\end{tabular} & 0.000969 \\ \hline
\end{tabular}
\end{center}
\end{table}

The second experiment was conducted to demonstrate the efficiency of the proposed algorithm. 1000 target points were randomly selected in the workspace of the planar 5-bar mechanism, the 6-UPS Stewart platform, and NRPM. 
P-FABRIK and the geometric approach were used to solve the IK, given the same target. The results for this experiment showed that the proposed algorithm has a similar computational efficiency to the geometric approach, as shown in Table \ref{tb:time-comp}.

\begin{table}[b]
\caption{Computational efficiency comparison between P-FABRIK and IK with geometric approach.}
\label{tb:time-comp}
\begin{center}
\begin{tabular}{cccc}
\hline
Mechanism                                                                                                 & Method                                                    & \begin{tabular}[c]{@{}c@{}}Average\\iterations\end{tabular} & \begin{tabular}[c]{@{}c@{}}Average time\\ {[}ms{]}\end{tabular} \\ \hline
\multicolumn{1}{c|}{\multirow{2}{*}{\begin{tabular}[c]{@{}c@{}}Planar 5-bar\\ mechanism\end{tabular}}} & P-FABRIK                                                     & 2.4                                                             & 0.0033                                                          \\ \cline{2-4} 
\multicolumn{1}{c|}{}                                                                                  & \begin{tabular}[c]{@{}c@{}}Geometric\\ approach\end{tabular} & 1.0                                                             & 0.0015                                                          \\ \hline
\multicolumn{1}{c|}{\multirow{2}{*}{\begin{tabular}[c]{@{}c@{}}6-UPS\\ Stewart platform\end{tabular}}} & P-FABRIK                                                     & 1.0                                                             & 0.0070                                                          \\ \cline{2-4} 
\multicolumn{1}{c|}{}                                                                                  & \begin{tabular}[c]{@{}c@{}}Geometric\\ approach\end{tabular} & 1.0                                                             & 0.0050                                                          \\ \hline
\multicolumn{1}{c|}{\multirow{2}{*}{NRPM}}                                                             & P-FABRIK                                                     & 1.9                                                             & 0.0548                                                          \\ \cline{2-4} 
\multicolumn{1}{c|}{}                                                                                  & \begin{tabular}[c]{@{}c@{}}Geometric\\ approach\end{tabular} & 1.0                                                             & 0.0717                                                          \\ \hline
\end{tabular}
\end{center}
\end{table}

The third experiment was conducted to demonstrate the robustness of the proposed algorithm in another trajectory-tracking scenario.
The target trajectory for the planar 5-bar mechanism was a circle whose center was at $(0$,$0$,$200)$, and its radius was $60$mm. A part of the upper arc of this trajectory exceeded its workspace. The target trajectory for the 6-UPS Stewart platform was a circle that was parallel to the $zoy$ surface. The center of the circle was at $(0$,$0$,$220)$, and its radius was $60$mm. A part of the upper arc of this trajectory exceeded its workspace. Its moving platform was parallel to the $xoy$ surface in this experiment. The target trajectory for NRPM was a circle that was parallel to the $zoy$ surface. The center of the circle was at $(0$,$0$,$200)$, and its radius was $100$mm. A part of the upper arc of this trajectory exceeded its workspace. Its moving platform was parallel to the $xoy$ surface in this experiment.

\begin{figure}[htbp]
  \centering
  \begin{subfigure}[t]{0.15\textwidth}
    \includegraphics[width=\textwidth]{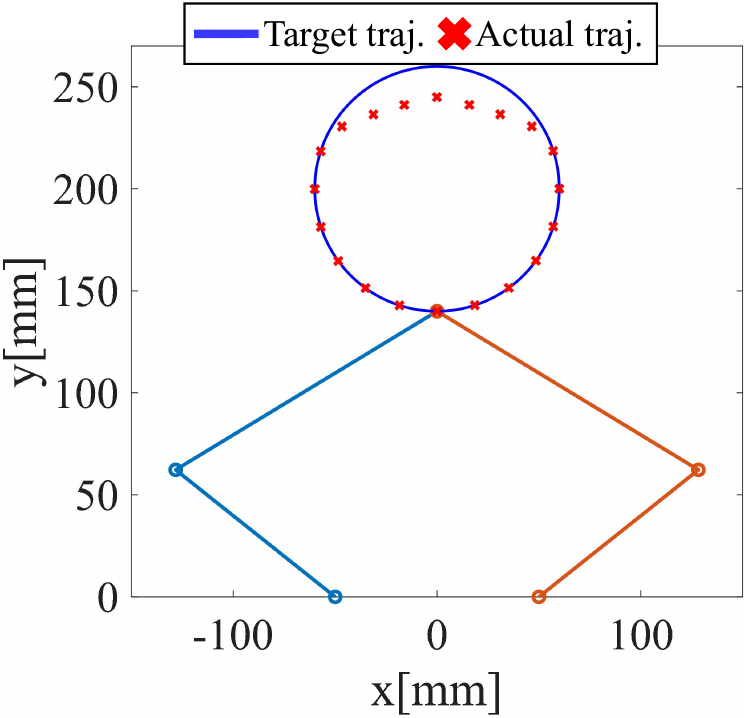}
    \caption{}
  \end{subfigure}
  \begin{subfigure}[t]{0.15\textwidth}
    \includegraphics[width=\textwidth]{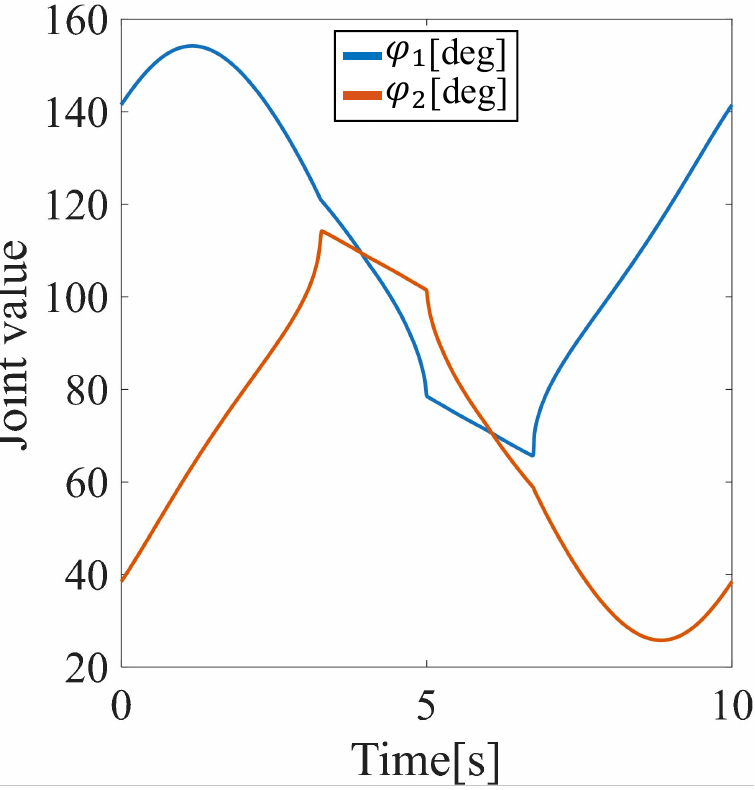}
    \caption{}
  \end{subfigure}
  \begin{subfigure}[t]{0.15\textwidth}
    \includegraphics[width=\textwidth]{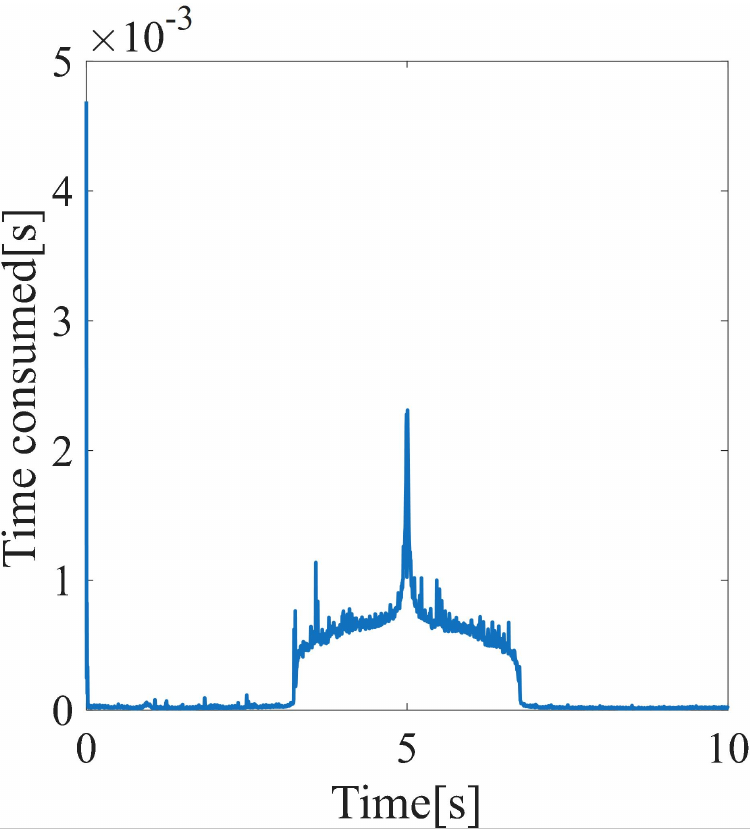}
    \caption{}
  \end{subfigure}
  \begin{subfigure}[t]{0.15\textwidth}
    \includegraphics[width=\textwidth]{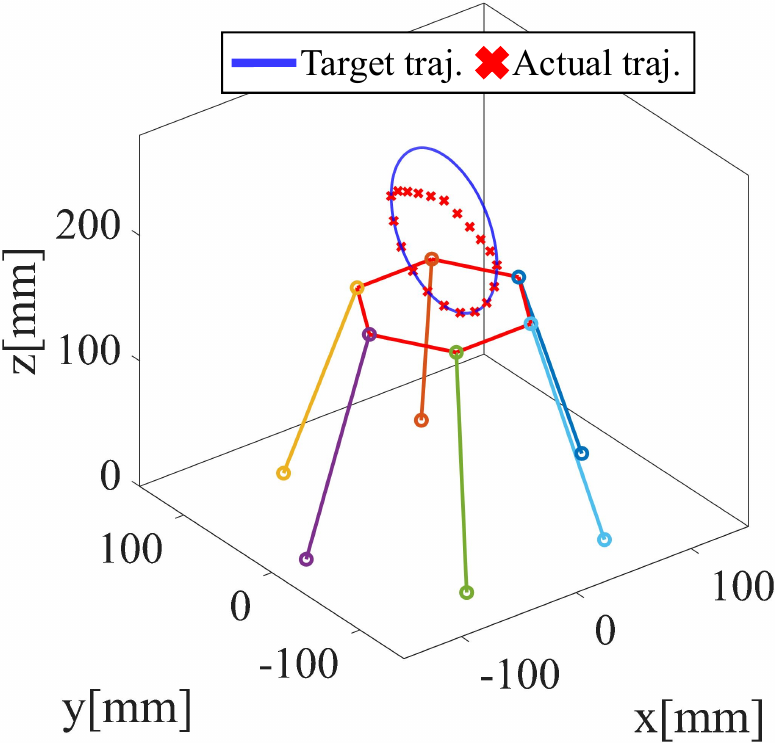}
    \caption{}
  \end{subfigure}
  \begin{subfigure}[t]{0.15\textwidth}
    \includegraphics[width=\textwidth]{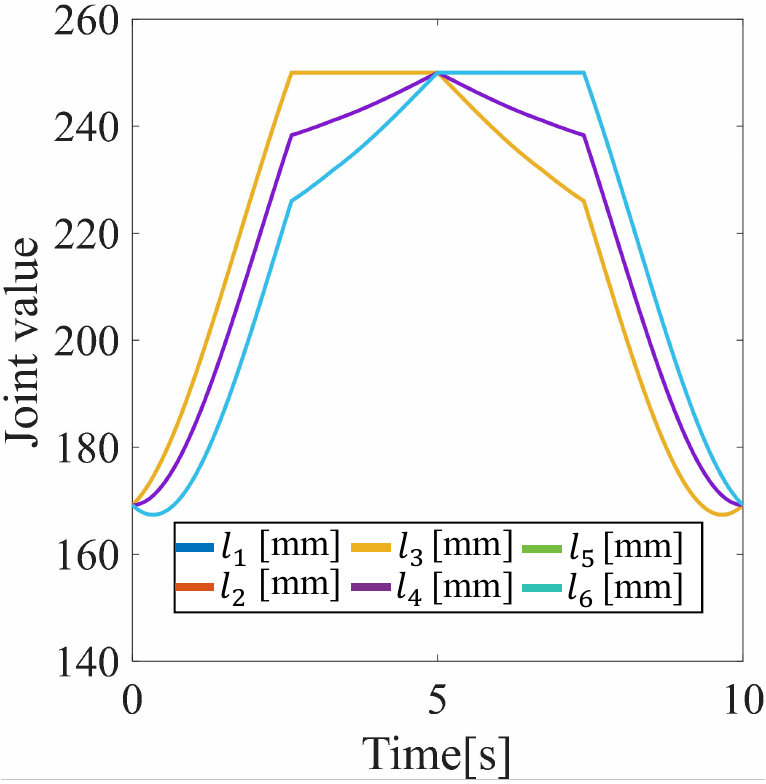}
    \caption{}
  \end{subfigure}
  \begin{subfigure}[t]{0.15\textwidth}
    \includegraphics[width=\textwidth]{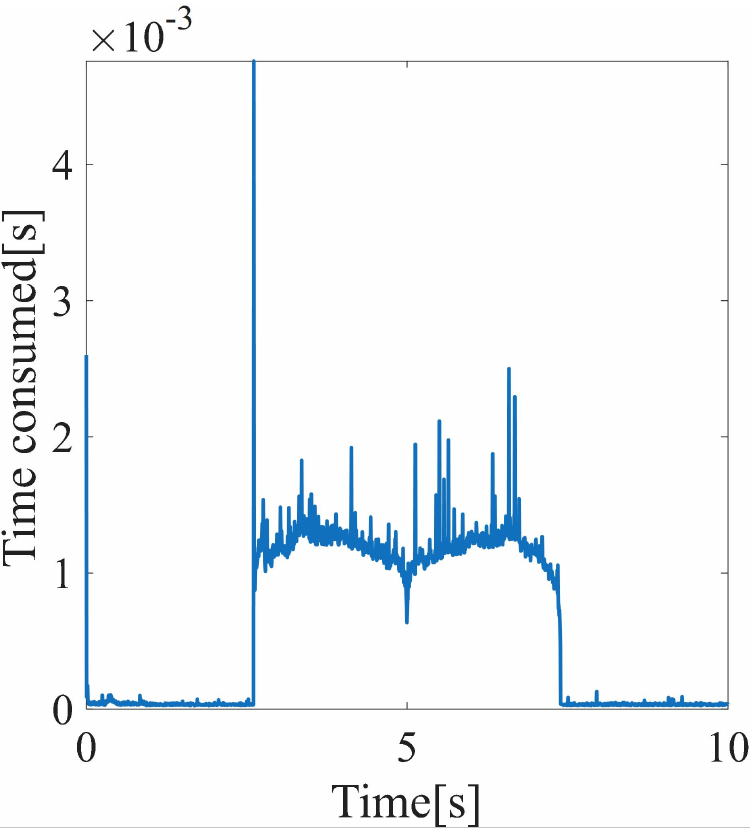}
    \caption{}
  \end{subfigure}
  \begin{subfigure}[t]{0.15\textwidth}
    \includegraphics[width=\textwidth]{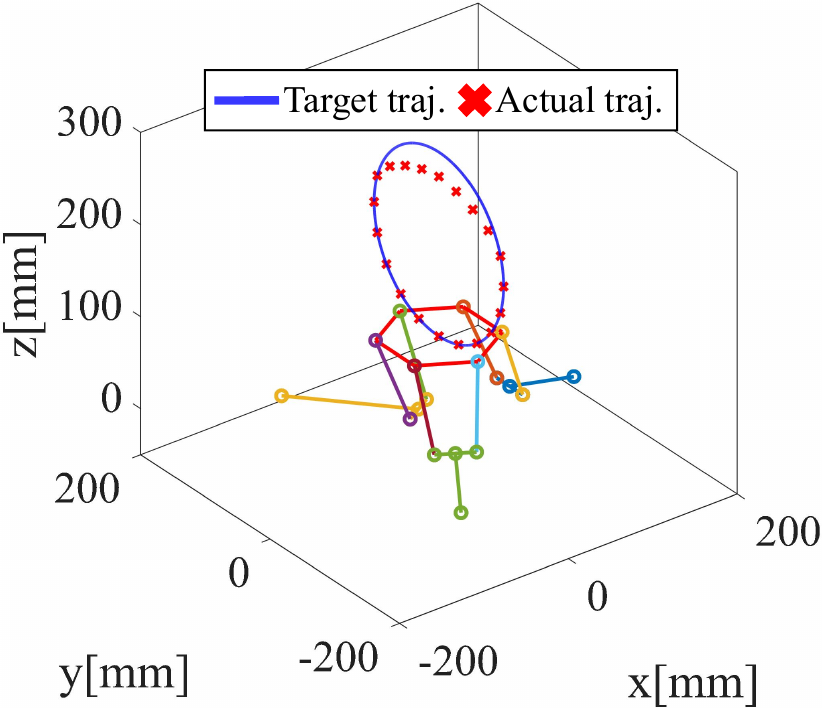}
    \caption{}
  \end{subfigure}
  \begin{subfigure}[t]{0.15\textwidth}
    \includegraphics[width=\textwidth]{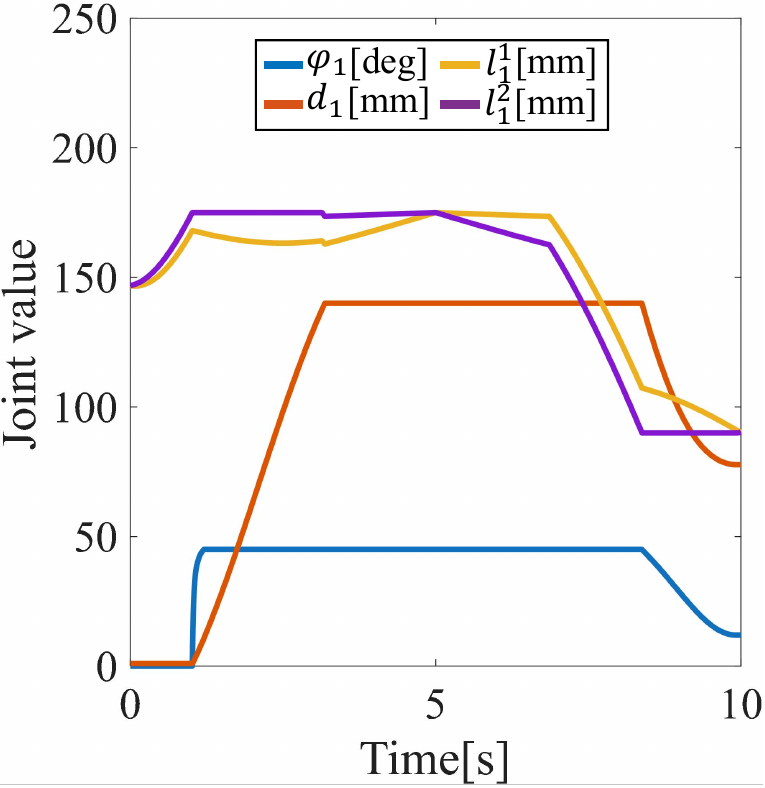}
    \caption{}
  \end{subfigure}
  \begin{subfigure}[t]{0.15\textwidth}
    \includegraphics[width=\textwidth]{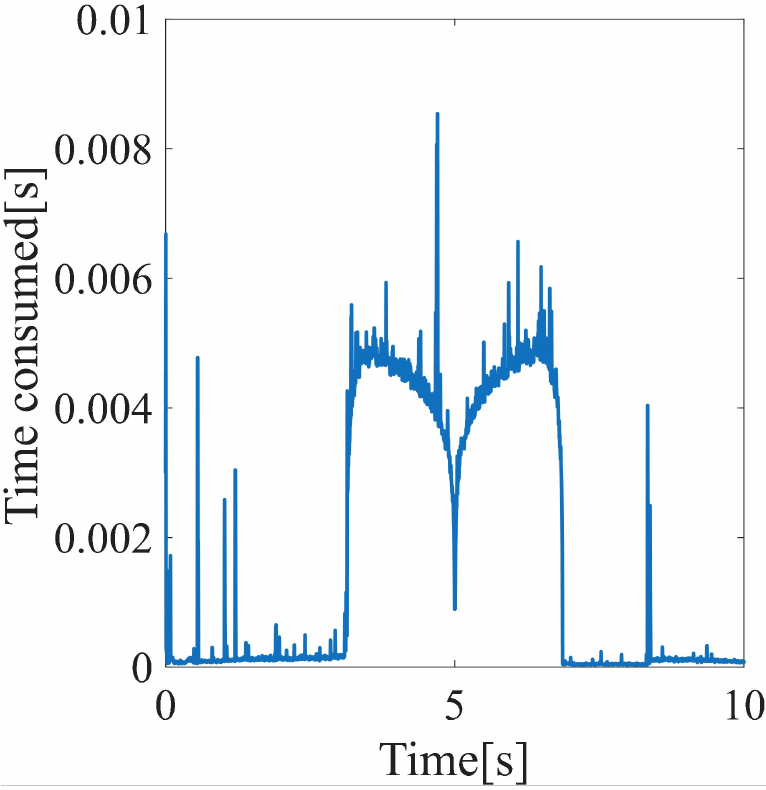}
    \caption{}
  \end{subfigure}
  \caption{Results of experiment on verifying the robustness of P-FABRIK. (a)-(c) The experiment results for the planar 5-bar mechanism. (d)-(f) The experiment results for the 6-UPS Stewart platform. (g)-(i) The experiment results for NRPM.}
  \label{fig:robust-exp}
\end{figure}

The result of this experiment was shown in Fig. \ref{fig:robust-exp}. 
The actual trajectory was an approximation to the target trajectory while remaining entirely within the workspace of each mechanism, as illustrated in Fig. \ref{fig:robust-exp}(a), (d), and (g). The joint values are presented in Fig. \ref{fig:robust-exp}(b), (e), and (h) to demonstrate the robustness of the proposed algorithm. To maintain clarity, Fig. \ref{fig:robust-exp}(h) illustrates the joint values for just sub-chain 1 of the NRPM, as the mechanism's symmetry and high joint number make displaying all values redundant.
The computational cost of trajectory tracking for each mechanism is presented in Fig. \ref{fig:robust-exp}(c), (f), and (i). When the target trajectory exceeds the mechanism's workspace, P-FABRIK requires more time to adjust the target with ATP adaptively, ensuring it lies within the feasible workspace.

The simulation experiments conducted in this section demonstrated the efficacy, computational efficiency, and robustness of the proposed P-FABRIK algorithm. 

\section{Conclusion}\label{sec:conslusions}
In this paper, a P-FABRIK algorithm was proposed as an inverse kinematics method to find one feasible solution for diverse
parallel mechanisms with computational efficiency, generality, and robustness.

The main idea of P-FABRIK was to decompose the parallel mechanism into several serial sub-chains first and then apply forward-reaching and backward-reaching stages on each sub-chain. The target for each end effector of the sub-chain was calculated via the geometric relations between the end effector and the controlled point on the moving platform. Target points that exceeded the workspace of the mechanism were also solvable using a self-adaptation algorithm,  which robustly provided an approximate IK solution.

The verification of P-FABRIK was presented with three distinct phases of parallel mechanisms, which include planar, standard, and redundant parallel mechanisms, showing its generality. Moreover, the implementation details were also demonstrated in the verification phases.

Three simulation experiments were conducted to demonstrate the proposed algorithm's efficacy, computational efficiency, and robustness to targets exceeded the workspace. The experiments involved the planar 5-bar mechanism, 6-UPS Stewart platform, and NRPM in the verification. A comparison was conducted between P-FABRIK and the geometric approach on all mechanisms to demonstrate a similar computational efficiency. Trajectory-tracking experiments were conducted on these mechanisms to verify the efficacy and robustness of P-FABRIK.
Future works can be conducted on extending the P-FABRIK algorithm to solve the forward kinematics of parallel mechanisms.



\bibliographystyle{IEEEtran}
\normalem
\balance
\bibliography{References}

\end{document}